\documentclass[final]{cvpr}

\usepackage{times}
\usepackage{epsfig}
\usepackage{graphicx}
\usepackage{amsmath}
\usepackage{amssymb}
\usepackage{multirow}
\usepackage{pifont}
\usepackage{amssymb}
\usepackage{xcolor}
\definecolor{linkc}{rgb}{0, 0.44, 0.74}
\definecolor{eqc}{rgb}{1, 0, 0}

\usepackage[pagebackref=false,breaklinks=true,letterpaper=true,colorlinks,urlcolor=linkc,citecolor=linkc,linkcolor=eqc,bookmarks=false]{hyperref}

\newcommand{\figdir}{figures}

\def\cvprPaperID{6703} 
\def\confYear{CVPR 2021}

\begin{document}

\title{Disentangled Cycle Consistency for Highly-realistic Virtual Try-On}
\author{Chongjian Ge$^1$ \quad
  Yibing Song$^{2\ast}$ \quad
  Yuying Ge$^1$ \quad
  Han Yang$^3$ \quad 
  Wei Liu$^4$ \quad 
  Ping Luo$^1$\\
  {$^1$The University of Hong Kong} \quad \quad {$^2$Tencent AI Lab} \\
  {$^3$ETH Zürich} \quad {$^4$Tencent Data Platform} \\
  \tt\small{\{rhettgee, yuyingge\}@connect.hku.hk \quad yibingsong.cv@gmail.com \quad hanyang@ethz.ch}\\
  \tt\small{wl2223@columbia.edu \quad pluo@cs.hku.hk}
 }
\maketitle

\pagestyle{empty}  
\thispagestyle{empty} 

\begin{abstract}
Image virtual try-on replaces the clothes on a person image with a desired in-shop clothes image. It is challenging because the person and the in-shop clothes are unpaired. Existing methods formulate virtual try-on as either in-painting or cycle consistency. Both of these two formulations encourage the generation networks to reconstruct the input image in a self-supervised manner. However, existing methods do not differentiate clothing and non-clothing regions. A straightforward generation impedes the virtual try-on quality because of the heavily coupled image contents. In this paper, we propose a Disentangled Cycle-consistency Try-On Network (DCTON). The DCTON is able to produce highly-realistic try-on images by disentangling important components of virtual try-on including clothes warping, skin synthesis, and image composition. 
Moreover, DCTON can be naturally trained in a  self-supervised manner following cycle consistency learning. Extensive experiments on challenging benchmarks show that DCTON outperforms state-of-the-art approaches favorably.
\end{abstract}

\begin{figure}[t]
\begin{center}
\begin{tabular}{c}
\includegraphics[width=1\linewidth]{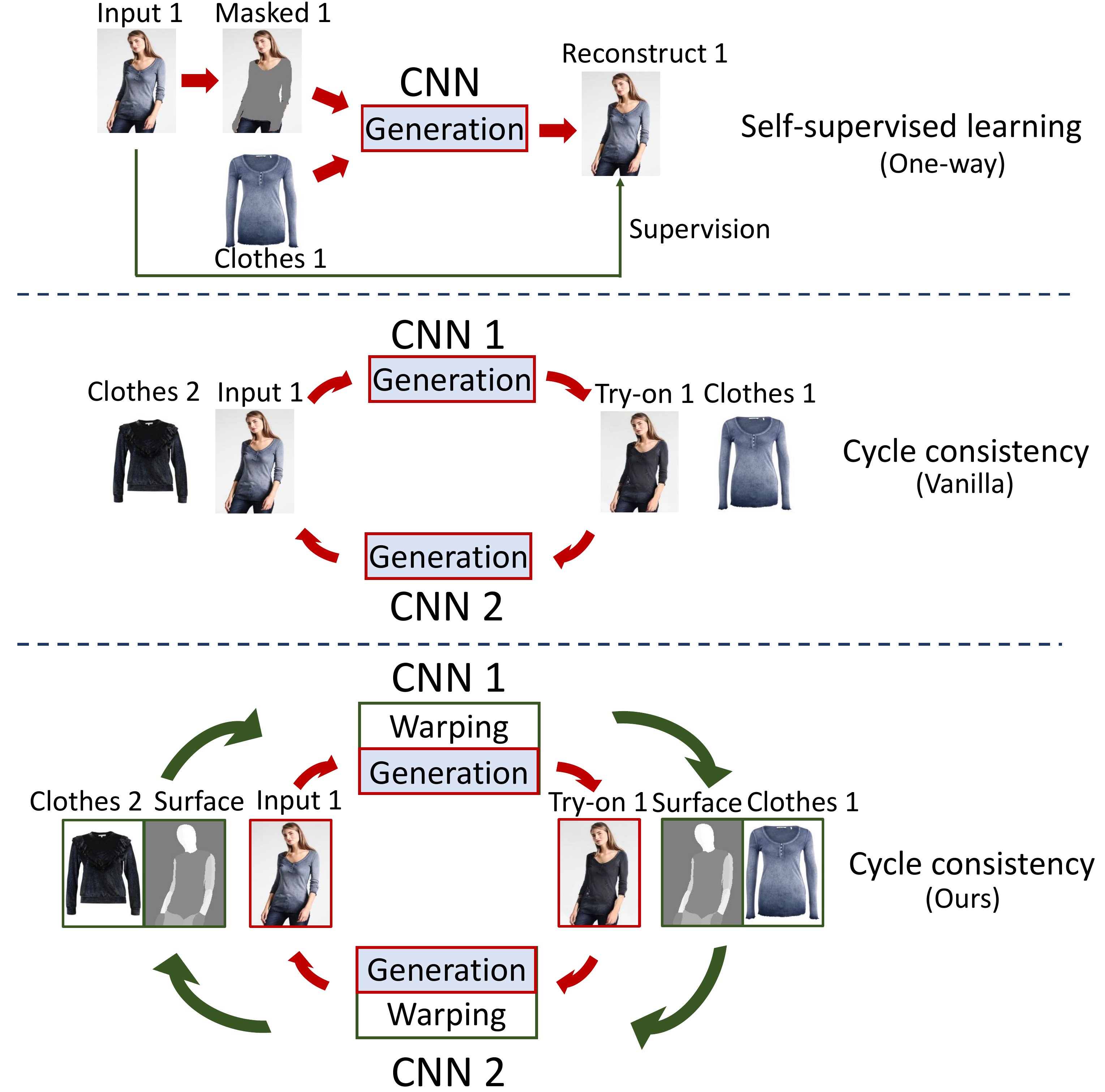}\\
\end{tabular}
\end{center}
\vspace{-2mm}
\caption{Comparison of virtual try-on pipelines. The inpainting methods (e.g., CP-VTON~\cite{cpvton} and ACGPN~\cite{ACGPN}) shown in the top row use one in-shop clothes to replace the same input clothes. The vanilla CycleGAN~\cite{cagan} shown in the middle row introduces two in-shop clothes for cycle consistency at the expense of generating coupled image contents (i.e., clothes, skin, and human poses). In the last row, we propose DCTON to disentangle virtual try-on as clothes warping and non-clothes generation, which is built upon vanilla cycle consistency for self-supervised learning.
}
\label{fig:teaser}
\end{figure}

\section{Introduction}
\renewcommand{\thefootnote}{\fnsymbol{footnote}}
\footnotetext[1]{Y. Song is the corresponding author. This work is done when C. Ge is an intern in Tencent AI Lab. The code is available at \url{https://github.com/ChongjianGE/DCTON}.}

Virtual try-on of fashion images aims at changing the clothes of a person with other in-shop clothes. There are wide applications including costume matching, fashion image editing, and clothes retrieval for e-commerce. Existing methods mainly focus on a direct try-on based on 2D images because of the available person images and in-shop clothes images online. However, these images are unpaired since the collection of images with multiple models, of which each model wears different and pixel-wise aligned clothes is infeasible.

\renewcommand{\tabcolsep}{1pt}
\def\swsix{0.14\linewidth}
\begin{figure*}[t]
    \small
	\begin{center}
		\begin{tabular}{cccccc}
		    \includegraphics[width=\swsix]{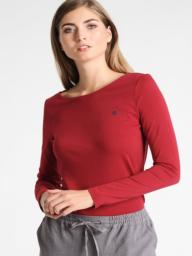}&
			\includegraphics[width=\swsix]{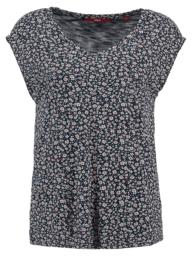}&
			\includegraphics[width=\swsix]{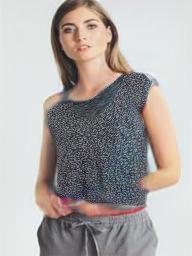}&
			\includegraphics[width=\swsix]{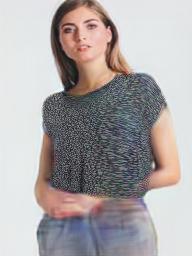}&
			\includegraphics[width=\swsix]{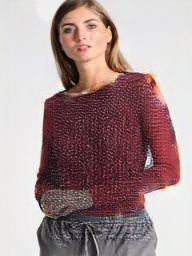}&
			\includegraphics[width=\swsix]{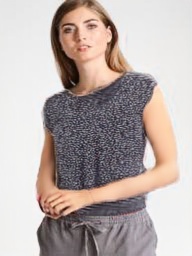}\\
			\includegraphics[width=\swsix]{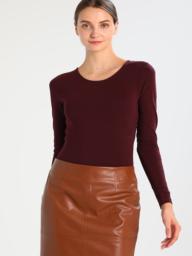}&
			\includegraphics[width=\swsix]{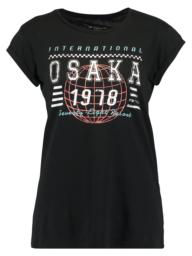}&
			\includegraphics[width=\swsix]{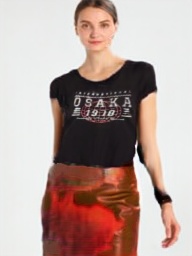}&
			\includegraphics[width=\swsix]{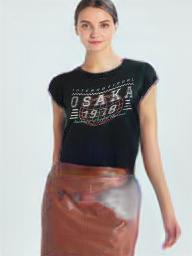}&
			\includegraphics[width=\swsix]{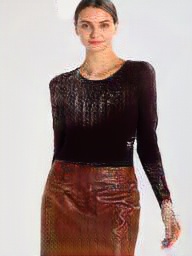}&
			\includegraphics[width=\swsix]{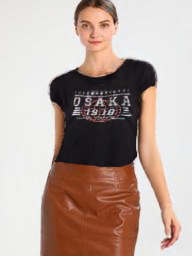}\\
			(a) Input&(b) Target& (c) ACGPN&(d) CP-VTON&(e) CA-GAN& (f) DCTON\\
		\end{tabular}
	\end{center}
	\vspace{-2mm}
	\caption{Virtual try-on comparisons. Inpainting based methods (ACGPN~\cite{ACGPN} and CP-VTON~\cite{cpvton}) are not effective to establish an accurate correspondence in (c) and (d) when the target clothes are significantly different from that in input images. Meanwhile, a heavily coupled content generation (CA-GAN~\cite{cagan}) brings salient artifacts as shown in (e). Different from existing methods, our DCTON disentangles virtual try-on as clothes warping, skin synthesis, and image composition in a cycle consistency training configuration. The network is learned to produce highly-realistic try-on results as shown in (f).}
	\label{fig:intro}
\end{figure*}

To handle unpaired images, existing methods such as VITON~\cite{viton}, CP-VTON~\cite{cpvton}, CP-VTON+~\cite{cpvton_plus}, and ACGPN~\cite{ACGPN} formulate virtual try-on as an inpainting problem. They first mask the clothes region of a person image, and then recover the clothes region by using the same in-shop clothes for  self-supervised network training. The pipeline is shown in the top row of  Fig.~\ref{fig:teaser}. It is regarded as a one-way reconstruction from the corrupted input image to its original image. Since these methods only use one clothes during training (i.e., clothes 1 is matched to input 1), they are not effective when the person image and the target in-shop clothes are significantly visually different. Examples are shown in Fig.~\ref{fig:intro}(c) and (d), where clothes with long sleeves will be changed to those with short sleeves. The arm region is not accurately generated as shown in the first row. Meanwhile, there are large artifacts on the skirts in the second row. Besides these observations, these methods utilize separate modules for virtual try-on such as thin plate splines (TPS)~\cite{tps} warping and semantic prediction. Their performance is limited due to a lack of end-to-end training for network potential exploitation.

Apart from the above inpainting-based methods, CA-GAN~\cite{cagan} incorporates cycle consistency for end-to-end network training. As shown in the middle row of Fig.~\ref{fig:teaser}, CA-GAN substitutes the clothes of an input person image (i.e., input 1) with an arbitrary target in-shop image (i.e., clothes 2). This network design improves correspondence matching between the person image and arbitrary target clothes. 
%
Nevertheless, it is still challenging to simultaneously generate the shape and the texture of clothes, the human skin, and the non-clothing contents in a cycle generative adversarial network (GAN). As shown in Fig.~\ref{fig:intro}(e), artifacts appear around the arms and the logo region. This indicates a straightforward generation via cycle consistency training is insufficient for high quality virtual try-on.

In this paper, we address aforementioned limitations by proposing a disentangled cycle-consistency try-on network (DCTON). It disentangles virtual try-on into three sub-modules. The first one is clothes warping module that preserves clothes design (e.g., collar style, sleeve cutting, and logo). The second one is skin synthesis module for occluded human body part generation (e.g., the arm of the blouse and vest in Fig.~\ref{fig:intro}). The third one is image composition module for output image generation.
During training, DCTON disentangles these three components from input images to constitute a try-on cycle for self-supervised learning.
Extensive experiments on the benchmark datasets show that DCTON performs favorably against state-of-the-art virtual try-on approaches.

\section{Related Work}
In this section, we review the literature of virtual try-on and cycle consistency for image generation. 

\subsection{Virtual Try-on}
Studies on virtual try-on derive from fashion editing~\cite{shi2019learning,han2019finet,zhu2017your,men2020controllable} for efficient clothes substitution. The computer graphics model~\cite{zhou2012image} and dimensionality reduction technique~\cite{ehara2006texture} are first developed for try-on generation. With the development of CNNs~\cite{song2017crest,song2018vital}, learning based methods evolve significantly. These methods can be categorized as 3D-based~\cite{drape,sekine2014virtual,pons2017clothcap,yang2016detailed} and 2D-based~\cite{cagan,viton,cpvton,ACGPN} methods. Due to the lightweight data collection, 2D methods suite real-world scenarios and thus become popular.

However, training a 2D-based try-on model is still challenging due to a lack of paired triplet data~\cite{viton,mpv} (i.e., a reference person, a target in-shop clothes, and the person wearing this clothes). Inspired by self-supervised learning, prior arts address this issue either in a one-way reconstruction~\cite{viton,cpvton,cpvton_plus,vtnfp,ACGPN,clothflow} or a vanilla cycle consistency generation~\cite{cagan}. For the one-way scheme, methods such as VITON~\cite{viton}, CP-VTON~\cite{cpvton} and CP-VTON+~\cite{cpvton_plus} first mask the region of both clothes and limbs, and then refill this region with either the same input clothes or the generated skin. These methods do not perform well when the target clothes is significantly different from that in the input images. Also, a lack of end-to-end training limit their generalization potential. 

The cycle consistency structure is employed in CA-GAN~\cite{cagan} for virtual try-on. By feeding the generator with shuffled training samples (i.e., the reference person and an arbitrary clothes), CA-GAN improves clothes characteristics preserving
while bringing undesirable artifacts in texture and body generation. This is because the generation of both clothes texture and occluded body parts is challenging for one network. To this end, our DCTON disentangles virtual try-on as clothes warping, skin synthesis, and image composition within a cycle consistency framework to produce highly-realistic try-on images.

\subsection{Cycle Consistency for Image Generation}
The self-supervised learning of cycle consistency introduces pixel-wise supervision for unpaired image-to-image generation~\cite{combogan,kim2017learning,taigman2016unsupervised,zhou2017genegan}.
In~\cite{cyclegan}, a CycleGAN framework is proposed for unpaired image synthesis.
The DualGAN is proposed in~\cite{dualgan} for image quality improvement. The relationships between different domains are explored in~\cite{kim2017learning} based on the cycle consistency. 

Cycle consistency learning has been applied to many applications including image style transfer~\cite{chang2018pairedcyclegan,song2017stylizing}, object tracking~\cite{wang2019unsupervised,wang2021unsupervised}, and photo enhancement~\cite{chen2018deep,yang2018image}. However, cycle consistency learning is not effective when handling person image generation~\cite{ma2017pose,siarohin2018deformable}, pose-guided animation~\cite{chan2019everybody}, image restoration~\cite{liu2020rethinking,wang2020rethinking}, and virtual try-on~\cite{viton}. Inspired by the cycle consistency scheme~\cite{cyclegan}, we reformulate the try-on task as a conditional unpaired image-to-image generation problem. The try-on result is conditionally generated by the images of the reference person and the target clothes. A straightforward cycle consistency is not effective for try-on as the generation of both clothes texture and occluded human parts is challenging. In this work, we disentangle try-on to several sub-modules for high-quality results production.


\begin{figure*}[t]
\begin{center}
\begin{tabular}{c}
\includegraphics[width=0.9\linewidth]{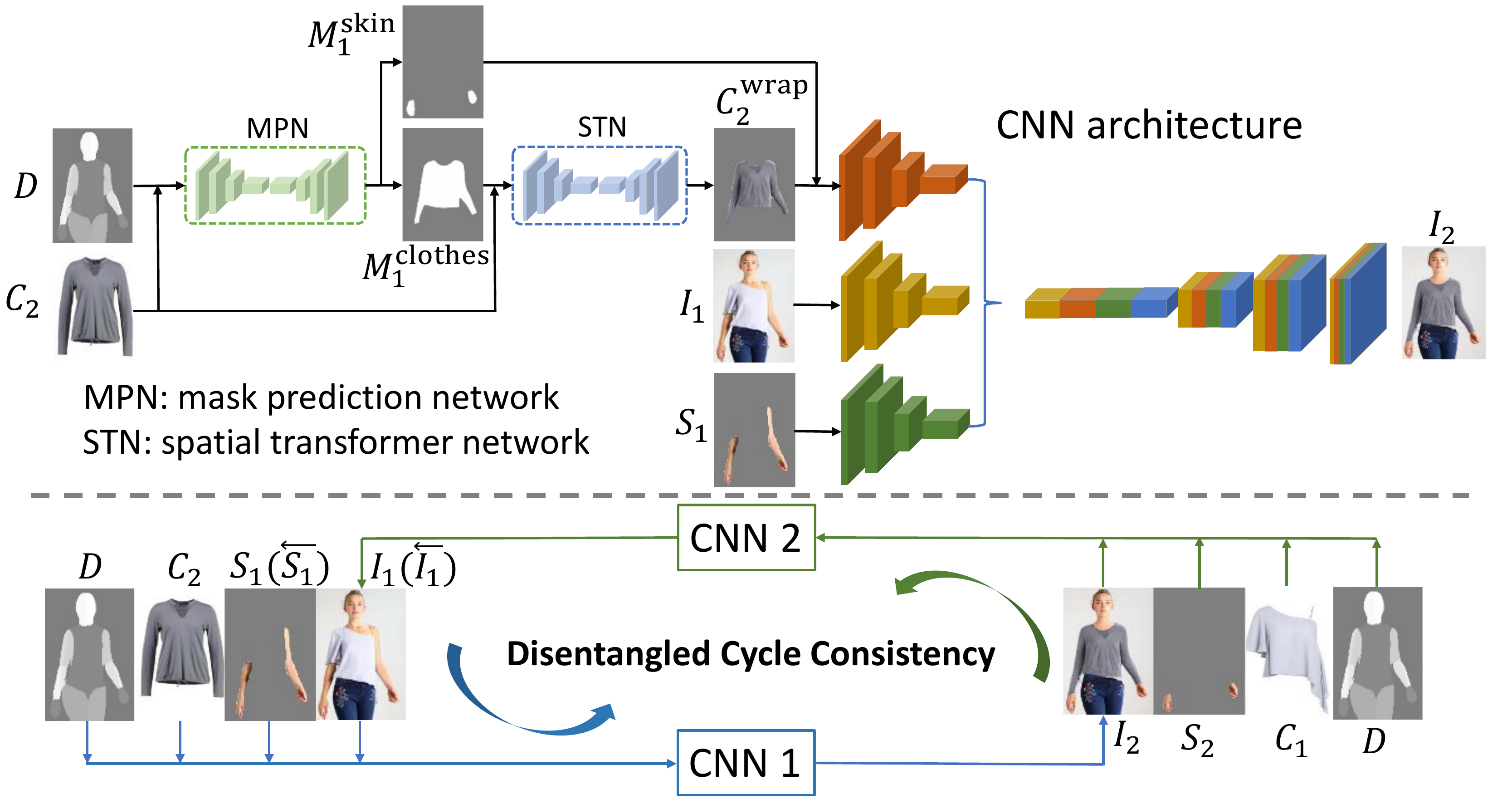}\\
\end{tabular}
\end{center}
\vspace{-2mm}
\caption{The pipeline of our disentangled cycle consistency framework. We show the CNN architecture above where there are clothes warping, skin synthesis, and image composition modules. The encoded features from these modules are concatenated to decode the output image. The cycle consistency is shown below where we use two CNNs with the same architecture. We send the output image from one CNN to another CNN as input to constitute self-supervision for end-to-end learning.} 
\label{fig:pipeline}
\end{figure*}

\section{Proposed Method}
We disentangle virtual try-on as clothes warping, skin synthesis, and image composition within the cycle consistency framework. Three encoders are utilized for the disentanglement. Fig.~\ref{fig:pipeline} shows an overview of our pipeline. In the following, we first illustrate each component of the disentanglement in Sec.~\ref{3.1}. Then, the cycle consistency training will be presented in Sec.~\ref{3.2} to empower networks for highly-realistic try-on generation.

\subsection{Disentangled Virtual Try-on} \label{3.1}
We use the subscript $1$ to illustrate the image contents related to the input clothes, and subscript $2$ to denote the image contents related to the target clothes. Specifically, we denote the input image as $I_1$, the in-shop target clothes image as $C_2$, the skin region of the input image as $S_1$, respectively. On the other hand, the in-shop clothes of the input image is denoted as $C_1$, the skin region of the output image is denoted as $S_2$, and the output image is denoted as $I_2$. These notations will be used to present the process of disentanglement.

\subsubsection{Clothes Warping}\label{sec:warp}
There are two sequentially-connected encoder-decoder networks and one encoder in the clothes warping module. We use the Densepose descriptor~\cite{densepose} to extract the human surface representation of the input image $I_1$, which is denoted as $D$. Then we send $D$ and $C_2$ into an encoder-decoder network named as MPN (mask prediction network). The MPN will produce the mask of the clothes region (i.e., $M_1^{\rm clothes}$) and skin region (i.e., $M_1^{\rm skin}$) of the input image, which are used as the prior guidance for further warping and generation, respectively. We train MPN with the supervision from the parsing labels of $I_1$ via the pixel-wise $L_1$ loss on each corresponding mask region. Note that different from previous works, we adopt the Densepose descriptor for human representation since it provides both the key point positions and semantic parsing results (e.g., body and arm shape), while vanilla 2D pose estimators can only provide the key point positions. The semantic parsing results improve our model to become sensitive around the human shapes for clothes fitting and characteristics generation.

After obtaining $M_1^{\rm clothes}$, we send it together with $C_2$ to the second encoder-decoder network, which is denoted as STN (spatial transformer network)~\cite{stn}. The STN will warp $C_2$ according to the guidance from $M_1^{\rm clothes}$. Specifically, STN first produces a transformation matrix $T$ and guides this matrix via Thin-Plate Spline (TPS)~\cite{tps} (i.e., $\mathcal{T}$) to warp the clothes image $C_2$. After obtaining the warped target clothes $C_2^{\rm warp}$ and the skin region $M_1^{\rm skin}$, we use an encoder to extract their pyramid features to further concatenate with other encoded features for output generation. The parameters of STN are kept fixed during the cycle consistency training. We pretrain the STN by only using the in-shop clothes image $C_1$ and the input image $I_1$. The loss function of STN can be written as:
\begin{equation}
    L_a=||\mathcal{T} (C_1)-I_1\odot M_1^{\rm clothes}||_1,
\end{equation}
where $M_1^{\rm clothes}$ is the mask region of the input image that is given by the parsing result, and $\odot$ is the element wise multiplication operation. 

Due to the huge variation of poses in the real-world try-on scenario, the original transformation matrix $T$ may not be effective enough to produce stable $\mathcal{T}$ during training. Simply adopting the STN is not capable of dealing with the large misalignment and deformation, thus bringing artifacts on the warped clothes $C_2^{\rm warp}$. We further incorporate a regularization term to robustly produce $T$. 
In practice, we first introduce a homography matrix $H$ to reduce the variations of $T$. For the $n$-th training iteration, we construct an objective function as:
\begin{equation}\label{eq:transf}
    R_b=||H\times T^{\rm n-1}-T^n||^2,
\end{equation}
where $T^{\rm n-1}$ is from the $(n-1)$-th iteration. We can use SVD~\cite{svd} to solve the Homogeneous Linear Least Squares problems as well as optimize $H$, and use the optimized $H$ to compute Eq.~\eqref{eq:transf} as a regularization term. As a result, the whole loss function to pretrain STN can be written as:
\begin{equation}
    L_{\rm STN}=L_a+R_b,
\end{equation}
where $R_b$ regularizes the transformation matrix $T$ during STN training. To this end, we have successfully disentangled the clothes warping via a sequential network.

\subsubsection{Skin Synthesis}\label{sec:skin}
The skin synthesis aims to recover the occluded human body regions during try-on. We extract the skin region of the input image (i.e., $S^1$) by using the input surface $D$. After obtaining $S_1$, another encoder branch is exploited  to capture its pyramid feature representations. The encoder we use contains the same architecture as that in Sec.~\ref{sec:warp}. The encoded features of $S^1$ are concatenated with other encoded features at each feature level to represent the output image $I_2$ in the CNN feature space.

\subsubsection{Image Composition}
After obtaining the encoded feature representations of warped clothes $C_2^{\rm warp}$ and skin image $S_1$, we send the input image $I_1$ into an encoder for global image representation. The encoder structure is the same as the other two encoders. We then concatenate the encoded features of $C_2^{\rm warp}$, $S_1$, and $I_1$ sequentially and send them into the decoders for output image $I_2$ generation. To this end, we perform clothes warping, skin synthesis, and image composition in three independent modules and fuse their feature representations to produce the try-on result.

\subsection{Cycle Consistency Training} \label{3.2}
Fig.~\ref{fig:pipeline} shows the cycle consistency construction. We generate a try-on result $I_2$ given an input image $I_1$ with its skin region $S_1$, target clothes $C_2$, and Densepose descriptor $D$.
In return, we use the generated try-on results $I_2$, the skin region brought by the target clothes $S_2$ (i.e., ${M_1^{\rm skin}}\odot I_2$), and the target clothes $C_1$ and $D$ as the input to generate an inversely predicted input image ${\mathop{I_1}\limits ^{\leftarrow}}$. Note that during the training process, the designed networks CNN$_1$ and CNN$_2$ in Fig.~\ref{fig:pipeline} share the same architectures. The cycle consistency will be established by enforcing ${\mathop{I_1}\limits ^{\leftarrow}} \approx {I_1}$ to formulate self-supervision.
We further illustrate the loss functions during the cycle consistency training as  follows:

{\flushleft \bf Adversarial Loss.} 
We introduce two discriminators $\mathcal{D}_p$ and $\mathcal{D}_s$ during the adversarial loss ${\mathcal{L}_{adv}}$ computation stage. The learned generators will synthesize a target try-on image $I_2$, an inversely predicted input image $\mathop{I_1}\limits ^{\leftarrow}$, a target skin image $S_2$, and an inversely predicted input skin image $\mathop{S_1}\limits ^{\leftarrow}$. We expect the appearance of both $\mathop{I_1}\limits ^{\leftarrow}$ and $I_2$ is similar to that of $I_1$, and the appearance of both $\mathop{S_1}\limits ^{\leftarrow}$ and $S_2$ is similar to that of $S_1$. The loss function can be written as follows:

\begin{equation}\label{eq:full_objective1}
\begin{aligned}
  {\mathcal{L}_{adv}}=&{\mathbb{E}_{{I_2},{\mathop{I_1}\limits ^{\leftarrow}}}}[log({D_p}(I_2)\cdot{D_p}({\mathop{I_1}\limits ^{\leftarrow}}))]+ \\
  &{\mathbb{E}_{{S_2},{\mathop{S_1}\limits ^{\leftarrow}}}}[log({D_s}({S_2})\cdot{D_s}({\mathop{S_1}\limits ^{\leftarrow}}))]+\\
  &\mathbb{E}_{{I_1},{{S}_1}}[log((1-D_p(I_1))\cdot(1-D_s({S}_1)))],
\end{aligned}
\end{equation} 
where ${\mathop{S_1}\limits ^{\leftarrow}}$ indicates the generated skin of ${\mathop{I_1}\limits ^{\leftarrow}}$.

{\flushleft \bf Cycle Consistency Loss.} In addition to the adversarial loss that ensures similar appearance distributions between the try-on results and the target images, we propose the cycle consistency loss to improve the pixel-wise self supervision. The cycle consistency loss term is based on ${\ell}_1$ on the synthesized try-on results and the corresponding skin regions, respectively. It can be written as follows:
\begin{equation}\label{eq:full_objective2}
\begin{aligned}
  {\mathcal{L}_{cyc}}=\big|\big| {{\mathop{I_1}\limits ^{\leftarrow}}}-{I_1}\big|\big|_{1}+\big|\big| {{\mathop{S_1}\limits ^{\leftarrow}}}- {S}_1\big|\big|_{1}.
\end{aligned}
\end{equation} 

{\flushleft \bf Content Preserving Loss.} For the contents within the human region excluding the skin and clothes regions, we aim to identically preserve them in the output try-on results. To this end, we design a content preserving loss term which measures the similarities between $I_1$ and $\mathop{I_1}\limits ^{\leftarrow}$, and $I_1$ and $I_2$ within this region. The loss term can be written as follows:
\begin{equation}\label{eq:full_objective3}
\begin{aligned}
  {\mathcal{L}_{pre}}=\big|\big| {M}\odot({I_2}- {I_1})\big|\big|_{1}+\big|\big| {M}\odot({{\mathop{I_1}\limits ^{\leftarrow}}}-{I_1})\big|\big|_{1},
\end{aligned}
\end{equation}
where $ M = 1-{{M_1^{\rm skin}}-{M_1^{\rm clothes}}}$ denotes the mask of the human body excluding the clothes and skin.

{\flushleft \bf Perceptual Loss.} We utilize the perceptual loss~\cite{vgg} to ensure similar CNN feature representations between the warped clothes. This improves the correspondence accuracy during clothes warping. The perceptual loss can be written as:
\begin{equation}\label{eq:full_objective4}
\begin{aligned}
  {\mathcal{L}_{vgg}}&=\sum_{l=1}\frac{1}{W_lH_lC_l}( \big|\big|\phi_l({{C_2}^{\rm wrap}}-{{M_1}^{\rm clothes}}\odot{I_2})\big|\big|_1\\
  &+\big|\big|\phi_l({{C_1}^{\rm wrap}}-{{M_2}^{\rm clothes}}\odot{\mathop{I_1}\limits ^{\leftarrow}})\big|\big|_1),
\end{aligned}
\end{equation}
where $\phi_l$ denotes the feature of the $l$-th layer in VGG19~\cite{perceptual}, and $W_l,H_l,C_l$ are the spatial parameters of the corresponding CNN features.

{\flushleft \bf Objective Function.}
Our final objective function consists of all the aforementioned loss terms and can be written as follows:
\begin{equation}\label{eq:full_objective}
  {\mathcal{L}_{all}}=\mathcal{L}_{adv}+\lambda_{cyc}\mathcal{L}_{cyc}+\lambda_{vgg}\mathcal{L}_{vgg}+\lambda_{pre}\mathcal{L}_{pre},
\end{equation}
where $\lambda_{cyc}$, $\lambda_{vgg}$ and $\lambda_{pre}$ are the constant scalars balancing the contributions from these loss terms.

\section{Experiments}

\begin{figure*}[t]
\begin{center}
\begin{tabular}{c}
\includegraphics[width=1.0\linewidth]{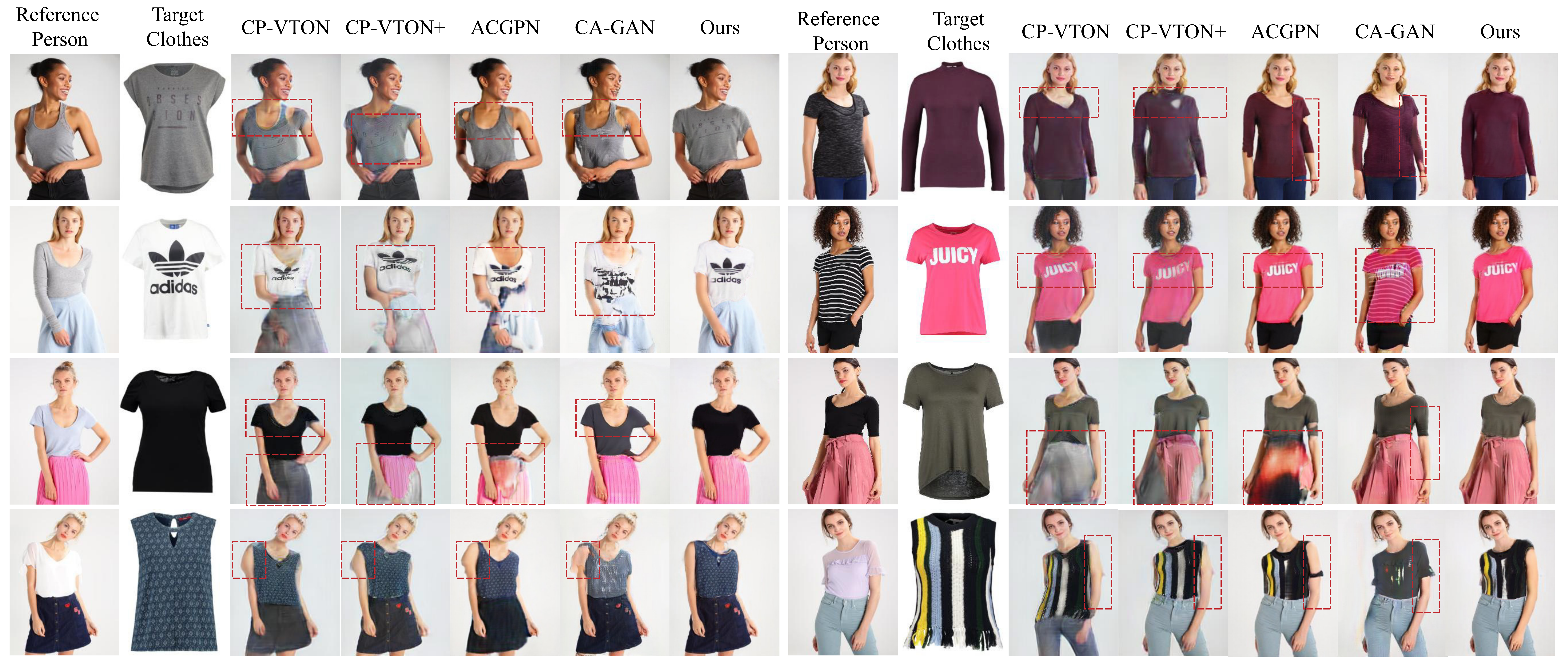}\\
\end{tabular}
\end{center}
\caption{Visual evaluation on the VITON dataset. Compared to existing methods~\cite{cpvton,cpvton_plus,ACGPN,cagan}, our DCTON is effective to preserve human body characteristics, and clothes textures, and generate occluded human body parts. These advantages enable DCTON to generate highly-realistic try-on results.}
\label{fig:visualcomparison}
\end{figure*}

In this section, we illustrate the benchmark datasets, implementation details, evaluation results, and ablation studies. The datasets we use are VITON and VITON-HD.

{\flushleft \bf VITON.} There are 19k image groups in this dataset. Each image group contains a frontal view of a model and an in-shop clothes image. We follow~\cite{cpvton} to exclude 2747 invalid image groups, and thus maintain a training set consisting of 14,221 groups and a testing set consisting of 2,032 groups. 

{\flushleft \bf VITON-HD.} The images in this dataset are the same to those of VITON but with a higher resolution of 512$\times$384. The VTION-HD dataset is more challenging since the results are in higher resolution where artifacts are more obvious on the try-on results.

\subsection{Implementation Details}

{\flushleft \bf Architectures.} Our network consists of four independent encoders, two decoders, and one pre-trained STN network. 
The architectures of the encoders and the decoders are from the Res-Unet~\cite{resunet}, and the corresponding discriminators are from PatchGAN~\cite{patchgan}. 
There are five convolutional layers with a stride number of 2 and two residual blocks in each encoder. The decoder in MPN and the decoder used to generate the try-on results both contain five deconvolutional layers.
The number of filters for the convolutional layers is 64, 128, 256, 512, 512 in each encoder, and 1536, 2048, 1024, 512, 256 in the final decoder used to output the try-on results, respectively. The STN 
is an encoder-decoder where the encoder consists of 5 convolutional layers with a stride number of 2. Each convolutional layer is followed by a max-pooling layer.

{\flushleft \bf Training and Testing.} We pretrain an STN network with paired data (i.e., the clothes region of the in-shop clothes image and the try-on results) by using the objective function in~\ref{3.1}. Then, we train DCTON end-to-end with the input of the model, the segmented skin, the Densepose descriptor, and the random in-shop clothes. DCTON is trained under 100 epochs. The parameter values of $\lambda_{cyc}$, $\lambda_{vgg}$, and $\lambda_{pre}$ are all set as 10. The initial learning rate is set to be 0.0002 and the model is optimized by the Adam optimizer with $\beta_1=0.5$ and $\beta_2=0.999$. During testing, we only use CNN$_2$ shown in Fig.~\ref{fig:pipeline} for online inference. The inputs to the network are the same as those during training.

\begin{figure*}[htb]
\begin{center}
\begin{tabular}{c}
\includegraphics[width=0.9\linewidth]{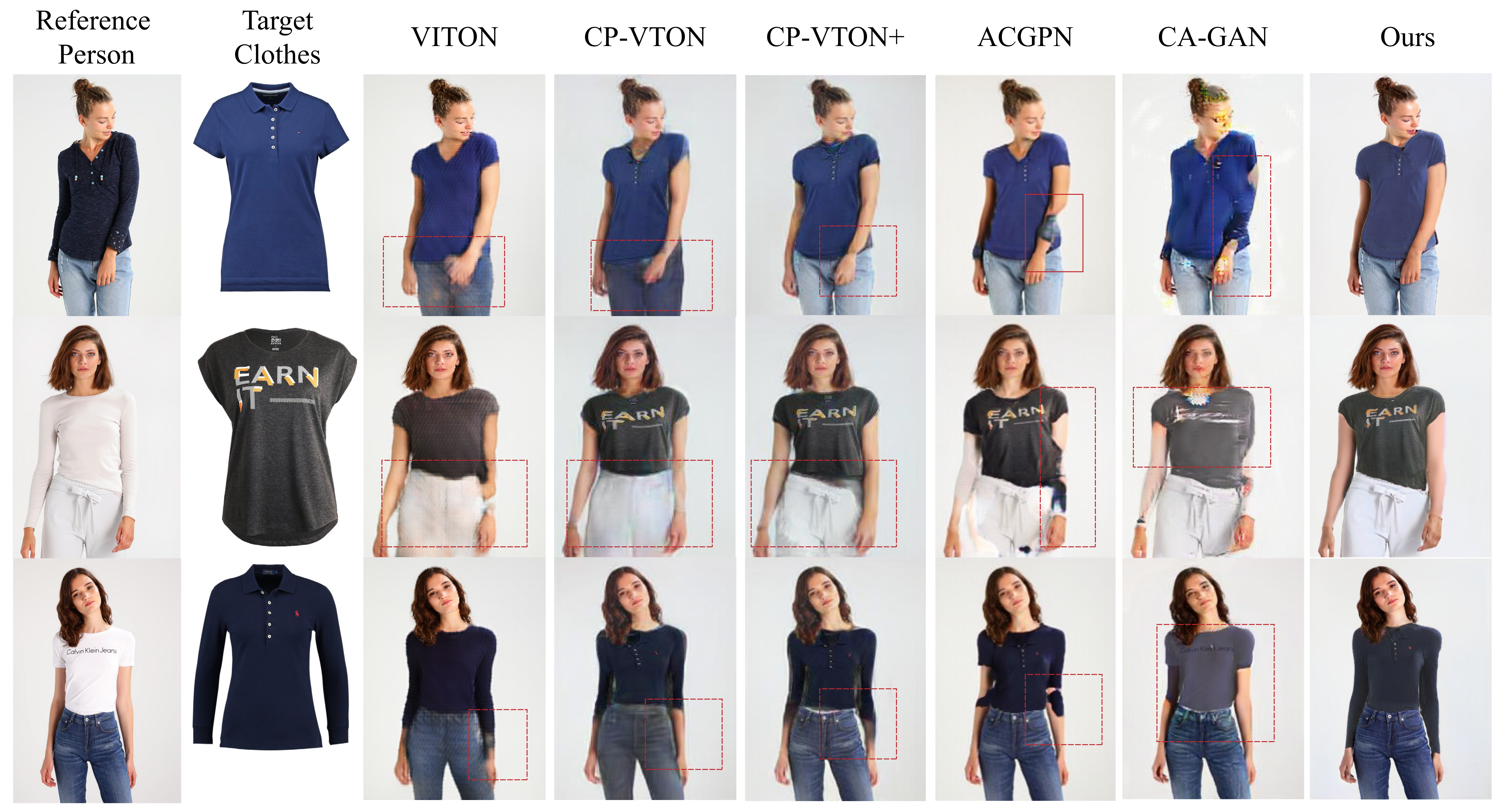}\\
\end{tabular}
\end{center}
\caption{Visual evaluations on the VITON-HD dataset. Our DCTON is effective to generate try-on results under a higher resolution. The results of existing methods are upsampled in this figure.}
\label{fig:vitonhq}
\end{figure*}


\subsection{Qualitative Evaluations}
We compare DCTON to inpainting based one-way reconstruction methods CP-VTON~\cite{cpvton}, CP-VTON+~\cite{cpvton_plus} and ACGPN~\cite{ACGPN}, and the vanilla cycle consistency method CA-GAN~\cite{cagan}. Fig.~\ref{fig:visualcomparison} shows the evaluation results. In the first row, we aim to indicate the clothes characteristics preserving ability of these methods. The target in-shop clothes and the input image clothes are significantly different. Existing methods do not attend to the target clothes and fit this clothes to the clothes region of the input image. To this end, limitations occur around the collar, sleeves, and the clothes boundaries.
%
%
These limitations are solved by our DCTON where the target in-shop clothes are arbitrary during training. We use various clothes to train DCTON with a high generalization ability.

In the second row, we aim to show the texture transfer ability of these methods. There are blur and distortions in the results generated by CP-VTON and CP-VTON+. Although these limitations are alleviated in the result of ACGPN, the whole clothes content is incorrectly generated. 
Compared with the vanilla cycle consistency method CA-GAN, DCTON is able to preserve the subtle embroiderer. Moreover, the subtle clothes texture is well preserved without distortion, due to the accurate clothes warping from STN.

In the third and last rows, we aim to indicate that whether existing methods maintain the non-clothes regions. The one-way inpainting methods are not effective for detail preservation (i.e., skirts in the third row). Moreover, there are limitations for these methods when generating occluded body parts including peculiar upper limbs, necks and hands.
%
From these examples, we conclude that the one-way inpainting methods bring blur on human bodies and clothes boundaries. They are not effective to preserve the target clothes characteristics (e.g., the collars and sleeves). This limitation is partially alleviated in CP-VTON+ and ACGPN. However, without using arbitrary clothes during training, the incorrect content generation around occluded human bodies occurs. The CAGAN uses cycle consistency to attend to clothes characteristics while the subtle textures are ignored. In comparison, we use disentangled cycle consistency during training. The learned DCTON is able to generate highly-realistic try-on results. The challenging factors including clothes textures warping, characteristics preserving, and occluded human body generation are effectively solved.

Besides evaluations on VITON, we show visual results on VITON-HD in Fig.~\ref{fig:vitonhq}. The VITON-HD dataset is more challenging for virtual try-on because the details are more obvious and artifacts are more salient. Nevertheless, our DCTON is effective to generate highly-realistic try-on results. Compared to existing methods, DCTON preserves the target clothes characteristics as shown around the collar regions in the first row. Meanwhile, DCTON is advantageous to generate occluded body parts (i.e., the arm region in the second row). Overall, our DCTON is effective for virtual try-on under such resolution where existing methods do not attempt. The results from existing methods in this figure are upsampled for a direct view comparison.


\subsection{Quantitative Evaluations} \label{quantitave}
We use the Fr\'{e}chet Inception Distance (FID)~\cite{fid} and Structural SIMilarity (SSIM)~\cite{ssim} metrics to measure the similarity of data distributions between the generated try-on results and the reference image (i.e., the reference person image). For a comprehensive comparison, Inception Score (IS)~\cite{is} is also utilized to measure the perceptual quality of synthesized images. To make the fair comparison, the quantitative results generated by different methods are evaluated under the same configurations.

\begin{table}[t]
\begin{center}
\small
\caption{The comparison of different methods under IS~\cite{is}, SSIM~\cite{ssim} and FID~\cite{fid} metrics. For IS and SSIM, the higher is the better. For FID, the lower is the better. DCTON$^\star$ denotes the DCTON without the skin synthesis encoder. And we use DCTON$^\diamond$ to indicate the DCTON without the regularization term in STN.}
\label{tab:comparison}
\vspace{0mm}
\renewcommand{\tabcolsep}{0.5mm}
\begin{tabular}{lcccc}
\hline
Methods & Dataset & IS~\cite{is}$\uparrow$  & SSIM~\cite{ssim}$\uparrow$  & FID~\cite{fid}$\downarrow$  \\
\hline
\hline
CA-GAN~\cite{cagan} & VITON & $2.56 \pm 0.09$ & 0.74 & 47.34 \\
VITON~\cite{viton} & VITON &$2.29\pm 0.07$ & 0.74 & 55.71 \\
CP-VTON~\cite{cpvton} & VITON &$2.59 \pm 0.13$ & 0.72 & 24.45 \\
CP-VTON+~\cite{cpvton_plus} & VITON &$2.75 \pm 0.14$ & 0.75 & 21.04 \\
ACGPN~\cite{ACGPN} & VITON &$2.69\pm 0.12$ & 0.81 & 16.64 \\
\hline
DCTON$^\star$ & VITON &$2.81 \pm 0.14$ & 0.74 & 18.12 \\
DCTON$^\diamond$ & VITON &$2.80 \pm 0.23$ & 0.79 & 15.70 \\
DCTON & VITON & \textbf{$\textbf{2.85} \pm \textbf{0.15}$} & \textbf{0.83} & \textbf{14.82} \\
\hline
DCTON & VITON-HD & \textbf{$\textbf{2.84} \pm \textbf{0.10}$} &\textbf{ 0.81} &\textbf{ 15.55} \\
\hline
\end{tabular}
\end{center}
\end{table}

Table \ref{tab:comparison} shows the SSIM, IS and FID scores by CA-GAN~\cite{cagan}, VITON~\cite{viton}, CP-VTON~\cite{cpvton}, CP-VTON+~\cite{cpvton_plus}, and ACGPN~\cite{ACGPN}. The IS results indicate that our DCTON outperforms CA-GAN, VITON, CP-VTON, CP-VTON+, and ACGPN by 0.29, 0.56, 0.26, 0.10, and 0.16, respectively. In the SSIM metric, our DCTON surpasses these methods by 0.09, 0.09, 0.11, 0.08, and 0.02, respectively. The lower FID score usually brings higher quality of the synthesized images. As such, our DCTON performs favorably against other methods. Note that even in the challenging VITON-HD dataset, our DCTON also brings considerable improvement. These results show the effectiveness and robustness of our method.

\begin{figure}[t]
\begin{center}
\begin{tabular}{c}
\includegraphics[width=\linewidth]{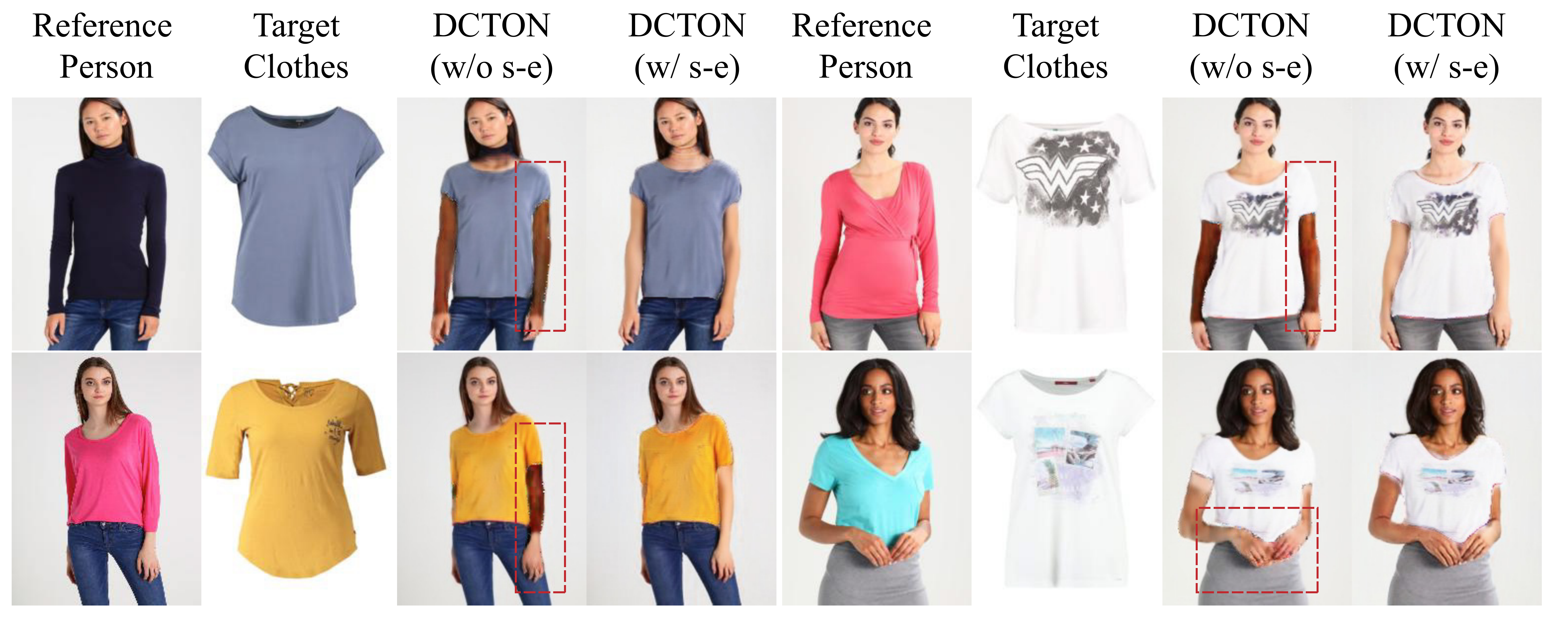}\\
\end{tabular}
\end{center}
\vspace{-2mm}
\caption{Ablation study on the effect of the skin synthesis encoder. S-e denotes the skin encoder. Without the prior features guidance provided by the skin synthesis encoder, DCTON$^\star$ is not capable of generating the realistic human skin.}
\label{fig:abla_skin}
\end{figure}

\begin{table}
\begin{center}
\caption{User study on the VITON test set. The ratio values indicate the percentages of subjects preferring DCTON.}
\vspace{-2mm}
\label{tab:userstudy}
\renewcommand{\tabcolsep}{1.2mm}
\begin{tabular}{cccccc}
\hline
Methods & CA~\cite{cagan} & VI~\cite{viton} & CP~\cite{cpvton} & CP+~\cite{cpvton_plus} & AC~\cite{ACGPN}  \\
\hline
\hline
DCTON  & 87.68\% & 80.32\% & 85.84\% & 79.82\% & 79.29\% \\
\hline
\end{tabular}
\end{center}
\end{table}

\begin{table}
\begin{center}
\setlength{\belowcaptionskip}{10pt}
\small
\caption{Time cost and computational complexity analysis.}
\vspace{-2mm}
\label{tab:comparison}
\renewcommand{\tabcolsep}{1.2mm}
\begin{tabular}{lccccc}
\hline
Methods & Dataset & Training Time  & \#Params  & FLOPS & FPS\\
\hline
\hline
ACGPN~\cite{ACGPN} & VITON & $\sim 40 h$  & 139M & 206G & 10 \\
\hline
DCTON & VITON & $\sim 44 h$   & 153M & \textbf{194G} & \textbf{19}\\
\hline
\end{tabular}
\end{center}
\end{table}

Besides the high-quality visual performance, DCTON is also advantageous by using less computational resource. We show the computational costs of ACPGN~\cite{ACGPN} and DCTON in Table~\ref{tab:comparison}. Under the same dataset (VITON) and hardware configurations (8 Nvidia Telsa V100 GPUs), the training time of DCTON is similar to that of ACGPN. Under only 1 V100 GPU, the online inference speed of DCTON is almost twice faster than that of ACPGN. We also analyze the model parameters and FLOPs in Table~\ref{tab:comparison}. DCTON contains more parameters while taking less FLOPs. The nearly real-time generation speed of DCTON (i.e., 19 FPS on 1 V100 GPU) is suitable for the online cloud service.

\subsection{User Study} 
The quantitative evaluation metrics are not sufficient to reflect the visual quality of the images as they measure the overall distributions of two image sets. To further evaluate the performance of existing methods, we conduct a user study where there are over 50 subjects. To make a fair comparison, 200 images from the VITON test set have been randomly selected for each method. A total of 1000 groups of generated images are provided for the user study on five comparing methods. The evaluation guidance is to consider the overall perceptual quality as well as fine-grained texture details. Each subject is randomly assigned with 100 image groups to select which result is better. Each image group contains a reference person, a target clothes, the generated results from DCTON, and another method for comparison. The results in Table~\ref{tab:userstudy} show that our DCTON achieves both higher perceptual quality and better texture details.

\begin{figure}[t]
\begin{center}
\begin{tabular}{c}
\includegraphics[width=\linewidth]{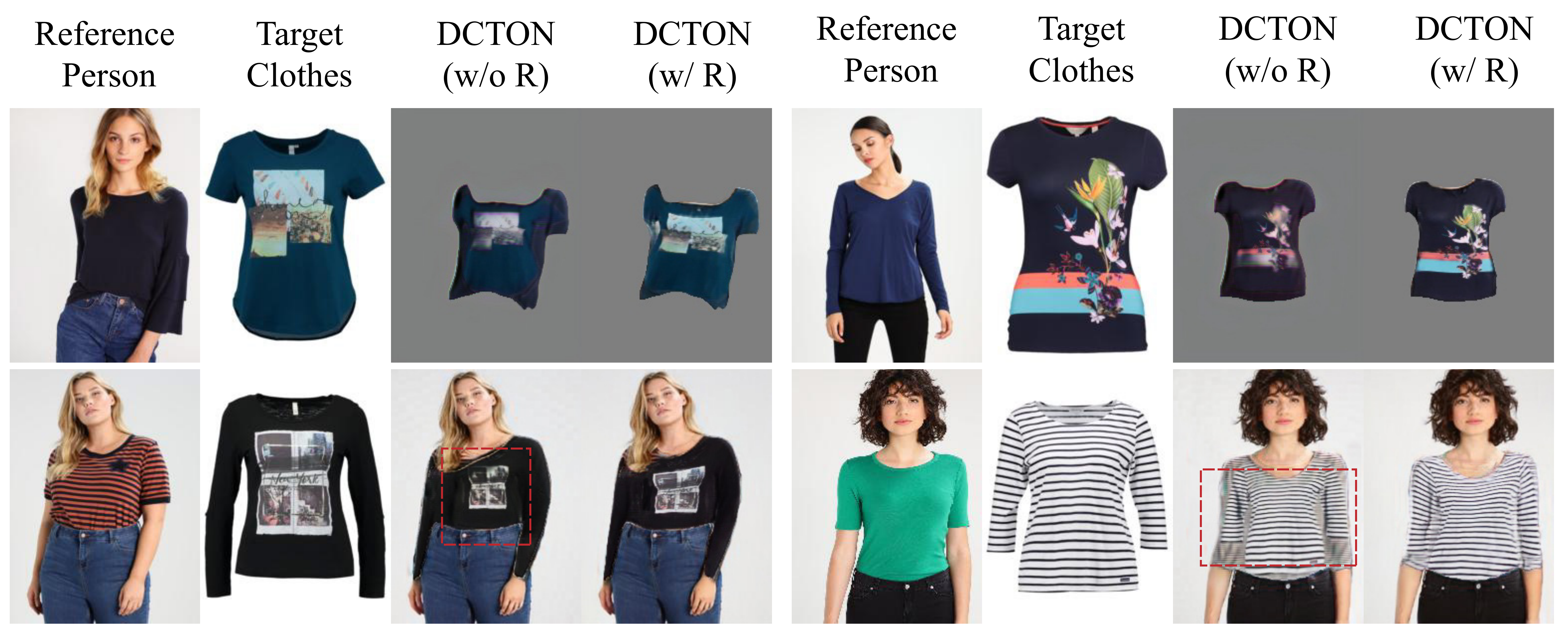}\\
\end{tabular}
\end{center}
\vspace{-2mm}
\caption{Ablation study on the effect of the proposed regularization term in STN. We denote R as the abbreviation of the regularization term. Without the regularization, STN will fail in warping the detailed textures.}
\label{fig:ablb_stn}
\vspace{-4mm}
\end{figure}

\subsection{Ablation Study}
We validate two components of DCTON (i.e., the generating module and warping module) in the ablation study. We use DCTON$^\star$ to indicate DCTON without the skin synthesis encoder, and DCTON$^\diamond$ to indicate DCTON without the regularization term in STN. We first assess the effects of the skin synthesis encoder. Quantitative results in Table~\ref{tab:comparison} show that after removing the skin encoder, the performance of DCTON$^\star$ will decrease but is still better than other methods by a margin. The visual comparison presented in Fig.~\ref{fig:abla_skin} shows that DCTON$^\star$ tends to generate the skin either with peculiar colors or blurring. An experiment is also performed to validate the proposed regularization term in STN. As shown in Fig.~\ref{fig:ablb_stn}, clothes with an obvious logos or embroiderer are presented as examples. From the first row in Fig.~\ref{fig:ablb_stn}, the STN module without the extra assistance of the proposed regularization term is prone to output the clothes with obvious distortion on the clothing texture. The second row in Fig.~\ref{fig:ablb_stn} shows that the regularization term facilitates STN to warp the target clothes in a proper manner.

\section{Concluding Remarks}
Virtual try-on methods typically consist of either a one-way reconstruction scheme or a vanilla cycle consistency configuration. However, limitations still exist when these methods generate photo-realistic try-on results. The one-way reconstruction scheme hinders existing methods from sufficient training, while the vanilla cycle consistency methods lack the texture preservation ability. In this paper, we proposed DCTON to disentangle virtual try-on as clothes warping, skin synthesis, and image composition. These modules are integrated within one framework for end-to-end cycle consistent training. Extensive experimental results validate that our DCTON achieves favorable performance compared to state-of-the-art virtual try-on approaches.

{\flushleft \bf Acknowledgement}. This work is supported by CCF-Tencent Open Fund and General Research Fund of HK No.27208720.  


\newpage
\clearpage

{\small
\bibliographystyle{ieee_fullname}
\bibliography{egbib}
}

\end{document}